\def\year{2016}\relax
\documentclass[letterpaper]{article}
\usepackage{aaai16}
\usepackage{times}
\usepackage{helvet}
\usepackage{courier}
\usepackage{graphicx}
\usepackage{amsmath}
\usepackage{amssymb}
\usepackage{multirow}
\usepackage{tabu}
\usepackage{wrapfig}

\newcommand{\bigpi}{\boldsymbol{\pi}}
\newcommand{\bigupsilon}{\boldsymbol{\upsilon}}
\newcommand{\bigrho}{\boldsymbol{\rho}}
\newcommand{\bigu}{\boldsymbol{u}}
\newcommand{\biglambda}{\boldsymbol{\lambda}}
\newcommand{\bigz}{\boldsymbol{z}}
\newcommand{\bigw}{\boldsymbol{w}}
\newcommand{\bigs}{\boldsymbol{s}}
\newcommand{\bigtheta}{\boldsymbol{\theta}}
\newcommand{\bigbeta}{\boldsymbol{\beta}}
\newcommand{\bigalpha}{\boldsymbol{\alpha}}
\newcommand{\biggamma}{\boldsymbol{\gamma}}
\newcommand{\bigb}{\boldsymbol{b}}
\newcommand{\bigt}{\boldsymbol{t}}

\begin{document}
\title{Jointly Modeling Topics and Intents with Global Order Structure}
\author{Bei Chen$^{\dagger}$, Jun Zhu$^{\dagger}$\thanks{Corresponding author.}, Nan Yang$^{\ddagger}$, Tian Tian$^{\dagger}$, Ming Zhou$^{\ddagger}$, Bo Zhang$^{\dagger}$\\
$^\dagger$Dept. of Comp. Sci. \& Tech., State Key Lab of Intell. Tech. \& Sys.,\\
Center for Bio-Inspired Computing Research, TNList, Tsinghua University, Beijing, 100084, China\\
$^\ddagger$Microsoft Research Asia, Beijing, 100080, China\\
\{chenbei12@mails., dcszj@, tiant13@mails., dcszb@\}tsinghua.edu.cn; \{nanya, mingzhou\}@microsoft.com\\
}

\maketitle
\begin{abstract}
Modeling document structure is of great importance for discourse analysis and related applications. The goal of this research is to capture the document intent structure by modeling documents as a mixture of topic words and rhetorical words. While the topics are relatively unchanged through one document, the rhetorical functions of sentences usually change following certain orders in discourse. We propose GMM-LDA, a topic modeling based Bayesian unsupervised model, to analyze the document intent structure cooperated with order information. Our model is flexible that has the ability to combine the annotations and do supervised learning. Additionally, entropic regularization can be introduced to model the significant divergence between topics and intents. We perform experiments in both unsupervised and supervised settings, results show the superiority of our model over several state-of-the-art baselines.
\end{abstract}

\section{Introduction}

People often organize utterances into meaningful and coherent documents, conforming to certain conventions and underlying structures. For example, scripts \cite{script14}, scientific papers \cite{rhetorical14}, official mails, news and encyclopedia articles all have relatively fixed discourse structure and exhibit recurrent patterns. Learning the document structure is of great importance for discourse analysis and has various applications, such as text generation \cite{prasad2005penn} and text summarization \cite{louis2010discourse}.

\begin{figure}[t]
\centering
\includegraphics[height=1.5in, width=3.0in]{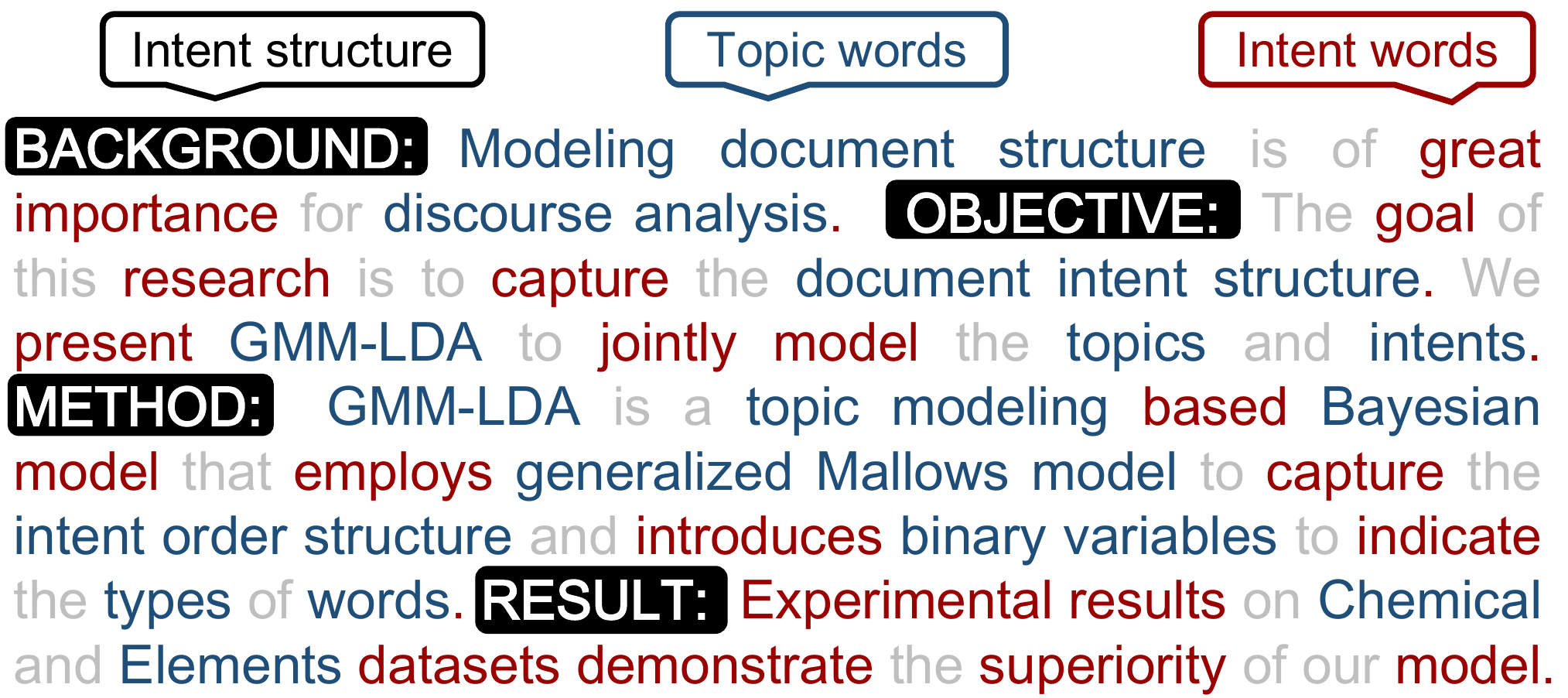}
\caption{Demonstration of the intent and topic words. Stop words (in gray) can be removed by preprocessing.}
\label{example}
\end{figure}

There are two important aspects of document structure learning: topic modeling and rhetorical structure modeling. Topic modeling assumes multiple topics often exist within a domain. It aims to discover the latent semantics of the documents, with many popular models such as Latent Dirichlet Allocation (LDA)~\cite{blei2003}, in which each document is posited as an admixture over an underlying set of topics, and each word is draw from a specific topic. Rhetorical structure modeling aims to uncover the underlying organization of documents. Inspired by the discourse theory~\cite{mann1988rhetorical}, each sentence in a document can be assigned a rhetorical function, or called \emph{intent}. For example, the sentences in a scientific paper may have different intents such as ``background'', ``objective'', ``method'' and ``result''. Fig~\ref{example} presents an example of the intent structure of an abstract.

Document-level topics and sentence-level intents usually show contradictory characteristics. For example, it is often sensible to assume that the topics are relatively unchanged through one document (e.g., in LDA), while the sentences' intents usually change following certain order in discourse. Furthermore, each document often follows a progression of nonrecurring coherent intents~\cite{hasan1976cohesion}, and the sentences with the same intent tend to appear within the same block of a document. Based on these observations, it's natural to think that jointly considering these two incompatible structures can help to model document better.

In this paper, we present a hierarchical Bayesian model to discover both the topic structure and rhetorical structure of documents by jointly considering topics and the above order structure in discourse. To this end, we assume that all the words can be divided into two types: \emph{topic word} and \emph{intent word}. Specifically, topic words in a document are relevant to the document's topic and spread throughout the document, while intent words mainly contribute to the rhetorical functions of the sentences. Following the example in Fig. \ref{example}, the words ``document'', ``discourse'' and ``intent'' are likely to be topic words; they indicate the specific research domain of this document. Meanwhile, the words ``result'', ``dataset'' and ``demonstrate'' are likely to be intent words and a sentence with these words may have the intent structure label ``result''.

We introduce a binary variable for each word to indicate its type, and model the topic structure and intent structure respectively using topic models. Inspired by the generalized Mallows model (GMM)~\cite{fligner1986distance}, we incorporate the intent order structure using GMM-Multi prior, which not only conforms to our intuition of nonrecurring coherent intents but also captures the global effects of the orders. To further improve the expressive ability, we present two important variants of GMM-LDA. One is to incorporate the known intent labels of sentences for supervised learning, and the other is to incorporate entropic regularization to better separate the words into two types under the regularized Bayesian inference framework~\cite{zhu2014bayesian}. Finally, we represent experiments on real datasets to demonstrate the effectiveness of our methods over several state-of-the-art baselines. To the best of our knowledge, we present the first model for topics and intents simultaneously, where the intent order structure is described globally by using GMM-Multi prior. In the rest paper, we first present GMM-LDA, followed by the supervised version and entropic regularization. Then we present experimental results with analysis. Finally, we discuss related work and conclude.

\section{Unsupervised GMM-LDA Model}

We consider the following document structure learning problem. We are given a corpus $\mathcal{D} = \{ \mathbf{s}_d\}_{d=1}^D$ with $D$ documents, where a document $\mathbf{s}_d$ is a sequence of $N_d$ sentences denoted by $\mathbf{s}_d = (\mathbf{w}_{d1}, \mathbf{w}_{d2}, ..., \mathbf{w}_{dN_d})$ and a sentence $\mathbf{w}_{ds}$ is a bag of $N_{ds}$ words denoted by $\mathbf{w}_{ds} = \{w_{dsm}\}_{m=1}^{N_{ds}}$. The size of the vocabulary is $V$. There are $T$ topics and $K$ intents in total. Topics are document-level, while intents are sentence-level. Our goal of document structure learning is to model the intent and the topic simultaneously and assign an intent label to each sentence in the corpus.

We build our models on the following assumptions, 1) {\bf{Type:}} Each word in the corpus is either an intent word or a topic word; 2) {\bf{Order:}} The intents of sentences within a document change following certain orders and the orders are similar within a domain; and 3) {\bf{Coherence:}} The same intent does not appear in disconnected portions of a document.

To characterize the type assumption, we associate each word with a binary variable to indicate whether it is an intent word or a topic word. For the order and coherence assumptions, we introduce a GMM-Multi prior to model the intent order structure. We start with the description of GMM-Multi prior, then GMM-LDA in detail with its inference method.

\subsection{GMM-Multi Prior for Intent Ordering}
According to the coherence assumption, the same intent could not be assigned to the unconnected portions of a document. To satisfy this assumption, we introduce a GMM-Multi prior over possible intent permutations. GMM-Multi, motivated by \cite{Chen09}, is an extension to the generalized Mallows model~\cite{fligner1986distance}. It concentrates probability mass on a small set of similar permutations around a canonical ordering $\bigpi_0$, which confirms to the intuition that the intent orders are similar within a domain. The \emph{inversion representation} of permutations is used instead of the direct order sequence: If we set the canonical ordering $\bigpi_0$ as the identity permutation $(1, 2, ..., K)$, then any permutation $\bigpi$ can be denoted as a $(K-1)$-dimensional vector $\bigupsilon = (\upsilon_1, \upsilon_2, ..., \upsilon_{K-1})$, where $\upsilon_k$ is the count of the numbers in $\bigpi$ that are before $k$ and greater than $k$. For instance, permutation $\bigpi = (2,1,4,5,3)$ can be represented as $\bigupsilon = (1, 0, 2, 0)$, where $\upsilon_3=2$ as there are two numbers, $4$ and $5$, that are before $3$ and greater than $3$. $\upsilon_K$ is omitted as it is always zero. Obviously, a one-to-one correlation exists between these two kinds of permutation representations.
Furthermore, each element of $\bigupsilon$ is independent of each other.
The marginal distribution over $\upsilon_k$ is
\begin{equation}\label{gmmk}
\mathrm{GMM}_k(\upsilon_k ; \rho_k) = \frac{e^{- \rho_k \upsilon_k}}{ \psi_k(\rho_k)},
\end{equation}where $\psi_k(\rho_k) = \frac{1-\exp{(-(K-k+1)\rho_k)}}{1-\exp{(-\rho_k)}}$ is the normalization factor and $\rho_k > 0$ is the dispersion parameter. Then the probability mass function with parameters $\bigrho = (\rho_1,\rho_2,...,\rho_{K-1}) $ can be written as the product over all the components, so that the number of parameters grows linearly with the number of intents. Due to the exponential form, the conjugate prior for $\rho_k$  is also an exponential distribution with two parameters :
\begin{equation}\label{gmm0}
\mathrm{GMM}_0(\rho_k ; \upsilon_{k,0},\nu_0) \propto e^{(-\rho_k \upsilon_{k,0} - \log\psi_k(\rho_k))\nu_0}.
\end{equation}Intuitively, $\nu_0$ is the number of previous trials and $\upsilon_{k,0}$ is the average number in previous trials for $\upsilon_k$. Let $\rho_0$ be the prior value for each $\rho_k$. By setting the maximum likelihood estimate of $\rho_k$ to be $\rho_0$, we can obtain $\upsilon_{k,0} = \frac{1}{\exp{(\rho_0)}-1} - \frac{K-k+1}{\exp{((K-k+1)\rho_0)}-1}$.

The GMM-Multi prior is defined over sentence intents $\bigz_d$ via a generative process on a given inversion representation of permutation $\bigupsilon_d$. Firstly, we draw a bag of intents $\bigu_{d}$ ($N_d$ elements) and each element $ u_{ds}\! \sim \! \mathrm{Multinomial} (\biglambda)$. Then $\sum_s\!{\mathbb{I}(u_{ds}=k)}$ represents the number of sentences of intent $k$ in this document. Secondly, we obtain the permutation of intents $\bigpi_d = \mathrm{Compute-}\pi(\bigupsilon_d)$, where $\mathrm{Compute-}\pi$ transforms $\bigupsilon_d$ into the intent permutation $\bigpi_d$. And finally, we obtain the intent structure of the sentences $\bigz_d =\mathrm{Compute-}z(\bigu_d, \bigpi_d)$, so that $z_{ds}$ is the intent label for sentence $\bigw_{ds}$. $\mathrm{Compute-}z$ is the algorithm to obtain the intent structure $\bigz_d$ using the bag of intents $\bigu_d$ and the permutation $\bigpi_d$. It arranges all the intent labels in $\bigu_d$ as the order of $\bigpi_d$ with the same intent labels appearing together. For example, when $\bigu_d = \{1,1,2,3,3,3,5\}$ and $\bigpi_d = (2,1,4,5,3)$, we can obtain $\bigz_d = (2,1,1,5,3,3,3)$. Note that not all the $K$ intent labels should appear in $\bigz_d$; it depends on the intent labels in $\bigu_d$. As in the example, $\bigu_d$ does not contain $4$, so $4$ does not appear in the intent structure $\bigz_d$.

\subsection{Generative Process of GMM-LDA}

\begin{figure}[t]
\centering
\includegraphics[height=1.7in, width=2.85in]{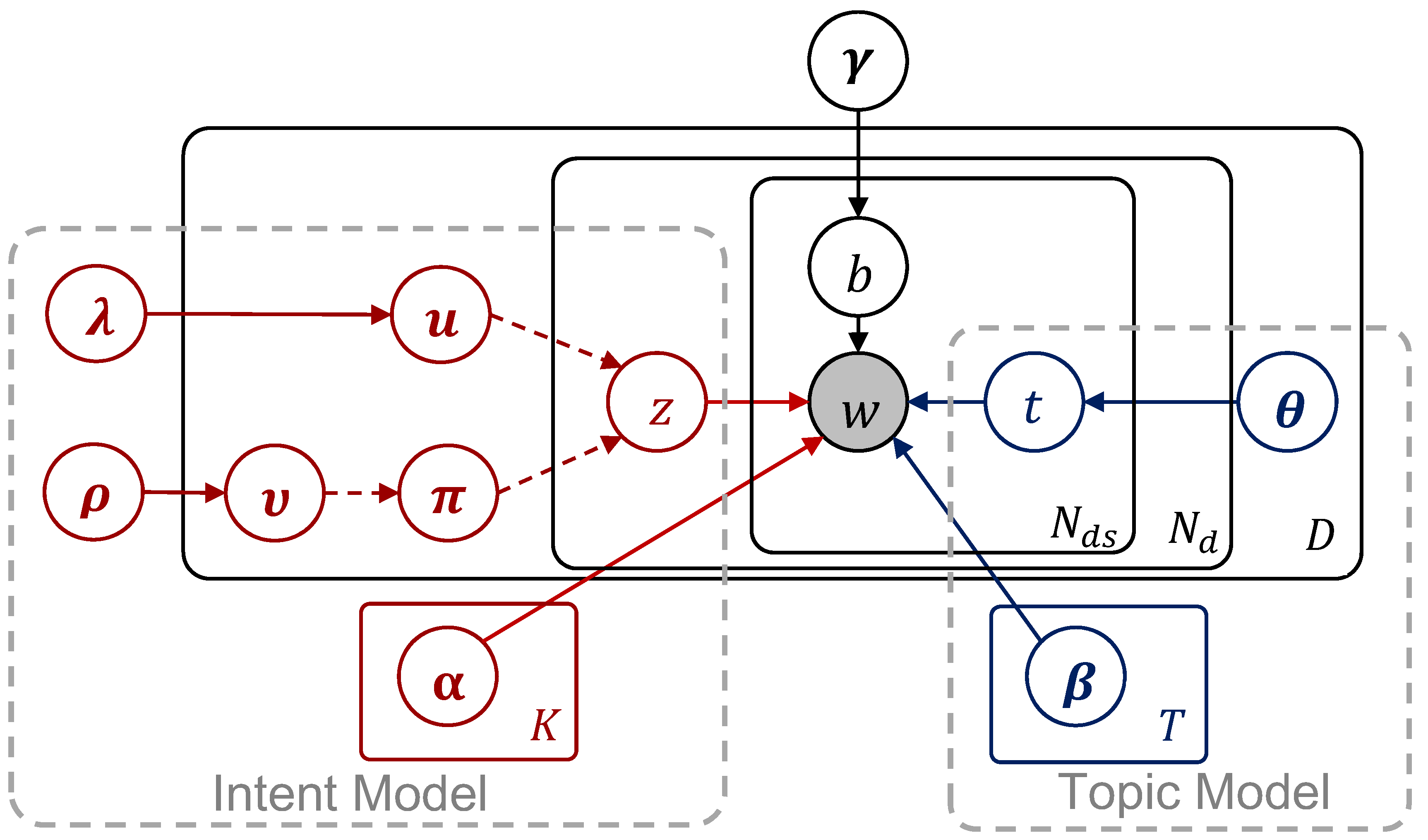}
\caption{The graphical structure of GMM-LDA.}
\label{figmodel}
\end{figure}

Now we present GMM-LDA, an unsupervised Bayesian generative model, as illustrated in Fig.~\ref{figmodel}. GMM-LDA simultaneously models the topics (the blue part in Fig.~\ref{figmodel}) and the intents (the red part in Fig.~\ref{figmodel}). The binary variable $b_{dsm}$ denotes the type of word $w_{dsm}$: if $b_{dsm}=1$, then $w_{dsm}$ is a topic word; if $b_{dsm}=0$, then $w_{dsm}$  is an intent word. Each sentence has a specific intent label $z_{ds}$, while each document $\bigs_d$ has a topic mixing distribution $\bigtheta_d$. For topic words, there is a document-specific topic model. It is a hierarchical Bayesian model that posits each document as an admixture of $T$ topics, where each topic $\bigbeta_{t}$ is a multinomial distribution over a $V$-word vocabulary, drawn from a language model $\bigbeta_t \sim \mathrm{Dirichlet} (\beta_0)$ ($t \in [T]$). For intent words, there is a rhetorical language model, in which the intent structure $\bigz_d$ of document $\bigs_d$ is generated from a bag of intent labels $\bigu_d$ and an intent permutation $\bigpi_d$ follows the GMM-Multi prior. The total number of intents is $K$, and each intent $\bigalpha_{k}$ is also a multinomial distribution over vocabulary, drawn from another language model $\bigalpha_k \sim \mathrm{Dirichlet} (\alpha_0)$ ($k \in [K]$).

For each document $\bigs_d$, the generating process is
\begin{enumerate}
\item Draw a topic proportion $\bigtheta_{d} \sim \mathrm{Dirichlet} (\theta_0)$.
\item Obtain the intent structure $\bigz_d \sim \mathrm{GMM}$-$\mathrm{Multi}(\bigrho,\biglambda)$, so that $z_{ds}$ is the intent label for sentence $\bigw_{ds}$.
\item For each word $w_{dsm}$ in document $d$,
\begin{enumerate}
\item Draw an indicator $b_{dsm} \sim \mathrm{Bernoulli} (\biggamma)$.
\item If $b_{dsm}=0$, then $w_{dsm}$ is from intent part: \\
draw $w_{dsm} \sim \mathrm{Multinomial} (\bigalpha_{z_{ds}})$.
\item if $b_{dsm}=1$, then $w_{dsm}$ is from topic part: \\
draw a topic $t_{dsm} \sim \mathrm{Multinomial} (\bigtheta_{d})$, and \\
draw $w_{dsm} \sim \mathrm{Multinomial} (\bigbeta_{t_{dsm}})$.
\end{enumerate}
\end{enumerate}
For fully-Bayesian GMM-LDA, we assume the following priors: $\biggamma \sim \mathrm{Beta} (\gamma_0)$, $\biglambda \sim \mathrm{Dirichlet} (\lambda_0)$, which is a distribution to express how likely each intent label is to appear regardless of positions, and each component of $\bigrho$ follows $\rho_k \sim \mathrm{GMM}_0(\rho_0,\nu_0), k \in [K-1]$. The variables subscripted with 0 are fixed prior hyperparameters.

\subsection{Collapsed Gibbs Sampling}

Let $\mathcal{M} = \{ \bigu, \bigrho, \bigupsilon, \bigt, \bigb \}$ denote all the variables to be learned during training. Then with Bayes' theorem, the posterior distribution of GMM-LDA is
\begin{equation}\label{bayes}
q(\mathcal{M} | \mathcal{D}) \propto p_0(\mathcal{M})p(\mathcal{D}|\mathcal{M}),
\end{equation}where $p_0(\mathcal{M})$ is the prior and $p(\mathcal{D}|\mathcal{M})$ is the likelihood. Then it can be learned with a collapsed Gibbs sampler due to the conjugate priors.  We naturally split the variables into four parts, namely $\bigu$, $\bigrho$, $\bigupsilon$ and $\{\bigt, \bigb \}$, then sample them using their posterior distributions respectively.

\textbf{For $\bigu$ : } For each document $\bigs_d$, $\bigu_d$ is a bag of intent labels with $N_d$ elements. We resample each element $u_{ds}$ via:
\begin{eqnarray*}\label{resampleu}
&&\!\!\!\!\!\!\!\!\!\!q(u_{ds} = x | \mathcal{D}, \mathcal{M}_{-u_{ds}})\propto p_0(u_{ds} = x)p(\bigs_d|\mathcal{M}, \mathcal{D}_{-\bigs_d}) \nonumber\\
\propto&&\!\!\!\!\!\!\!\!\!\! (f_{u=x}^{-ds} + \lambda_0)
\prod\limits_{k=1}^K \frac{\Gamma(f_{0,k,\cdot}^{-d}+V\alpha_0)}{\Gamma(f_{0,k,\cdot}+V\alpha_0)}
\prod\limits_{v=1}^V \frac{\Gamma(f_{0,k,v}+\alpha_0)}{\Gamma(f_{0,k,v}^{-d}+\alpha_0)},
\end{eqnarray*}where the subscript ``$-$" denotes that some elements are omitted from a set, e.g., $\mathcal{M}_{-u_{ds}}$ is the set $\mathcal{M}$ except $u_{ds}$ and $\mathcal{D}_{-\bigs_d}$ is the set of all documents in $\mathcal{D}$ except $\bigs_d$. Let $\bigu=\{\bigu_d\}_{d=1}^D$ denote all the intent labels in the corpus. $f_{u=x}^{-ds}$ is the count of times that intent label $x$ appears in $\bigu$ except $u_{ds}$. Let $w_v$ denote the $v$-th word in the vocabulary\footnote{$w_v$ is different from $w_{dsm}$.  $w_{dsm}$ is a word in sentence $\bigw_{ds}$ .}, where $v \in [V]$. $f_{0,k,v}$ counts the times that $w_v$ appears as an intent word in the sentences with intent label $k$. Then, $f_{0,k,\cdot} = \sum_{v=1}^V f_{0,k,v}$. The superscript ``${-d}$'' indicates that the frequency is calculated over all documents except $\bigs_d$.

\textbf{For $\bigrho$ : } We update each $\rho_k$ from its posterior distribution:
\begin{eqnarray}
q(\rho_k| \mathcal{D}, \!\mathcal{M}_{-\rho_k}\!)\! =\! \mathrm{GMM}_0 \Big( \rho_k ;\!\!
\frac{\sum_{d=1}^D \!\!\upsilon_{d,k}\!+\!\upsilon_{k,0}\nu_0}{D\!+\!\nu_0},D\!+\!\nu_0 \Big),\!\!\!\!\!\!\!\!\!\!\!\!\!\!\nonumber
\end{eqnarray}Since the normalization constant is unknown, it is intractable to sample directly from $\mathrm{GMM}_0$. Fortunately, slice sampling~\cite{neal2003slice} can be used to solve this problem.

\textbf{For $\bigupsilon$ : } For the inversion representation $\bigupsilon_d$ of document $\bigs_d$, each $\upsilon_{d,k}$ can be resampled independently from its posterior distribution:
\begin{eqnarray*}\label{resampleup}
&&\!\!\!\!\!\!\!\!\!\!q(\upsilon_{d,k} = v | \mathcal{D}, \mathcal{M}_{-\upsilon_{d,k}})\!\propto\! p_0(\upsilon_{d,k} = v)p(\bigs_d|\mathcal{M}, \mathcal{D}_{-\bigs_d}) \nonumber\\
\propto&& \!\!\!\!\!\!\!\!\!\!p_0(\upsilon_{d,k} = v)
\prod\limits_{k=1}^K \frac{\Gamma(f_{0,k,\cdot}^{-d}+V\alpha_0)}{\Gamma(f_{0,k,\cdot}+V\alpha_0)}
\prod\limits_{v=1}^V \frac{\Gamma(f_{0,k,v}+\alpha_0)}{\Gamma(f_{0,k,v}^{-d}+\alpha_0)},
\end{eqnarray*}where $p_0(\upsilon_{d,k} = v)=\mathrm{GMM}_k(\upsilon_{d,k} = v;\rho_k)$ is the prior.

\textbf{For $\bigt, \bigb $ : } Since $t_{dsm}$ has meaningful value only when $b_{dsm}=1$, we jointly sample $b_{dsm}$ and $t_{dsm}$ for the topics. The joint distributions are
\begin{eqnarray*}\label{resampletb}
&&q(b_{dsm}=1,t_{dsm}=t | \mathcal{D}, \mathcal{M}_{-\{b_{dsm},t_{dsm}\}})\nonumber\\
\propto&& (f_{b=1}^{-dsm}+\gamma_0) \frac{f_{1,t,v}^{-dsm}+\beta_0}{f_{1,t,\cdot}^{-dsm}+V\beta_0}
\frac{f_{1,d,t}^{-dsm}+\theta_0}{f_{1,d,\cdot}^{-dsm}+T\theta_0} ,\nonumber\\
&&q(b_{dsm}=0,t_{dsm}=\emptyset | \mathcal{D}, \mathcal{M}_{-\{b_{dsm},t_{dsm}\}})\nonumber\\
\propto&& (f_{b=0}^{-dsm}+\gamma_0) \frac{f_{0,z_{ds},v}^{-dsm}+\alpha_0}{f_{0,z_{ds},\cdot}^{-dsm}+V\alpha_0},
\end{eqnarray*}where $f_{b=1}$ is the number of topic words in the corpus and $f_{b=0}$ is the number of intent words. $f_{1,t,v}$ counts all the times that $w_v$ appears in the corpus with indicator variable value $1$ and topic label $t$. Then, $f_{1,t,\cdot}\!=\!\!\sum_{v=1}^V f_{1,t,v}$. $f_{1,d,t}$ is the number of words in document $\bigs_d$ with indicator variable value $1$ and topic label $t$. Then $f_{1,d,\cdot}=\sum_{t=1}^T f_{1,d,t}$ . The superscript ``${-dsm}$'' indicates that the frequency is calculated except $w_{dsm}$. According to the joint distribution of $b_{dsm}$ and $t_{dsm}$, we compute the $T+1$ non-zero probabilities, then do normalization and sample from the resulting multinomial.

\section{Supervised GMM-LDA}

GMM-LDA is unsupervised; it learns the document structure without human annotation. As suggested by existing supervised topic models~\cite{blei2010supervised,wang2009simultaneous,zhu2012medlda}, the predictive power can be greatly improved with a small amount of labeled documents. Here, we present a supervised GMM-LDA which combines the known intent labels of sentences during learning.

We consider the setting where a part of the documents in the corpus are labeled, that is, each sentence is assigned an intent label. Our goal is to develop a supervised model to learn the intent structures for the remaining unlabeled documents. The simplest way to leverage the label information is to use the labels directly during learning instead of sampling them. However, the GMM describes the intent order structure in a global way, which makes the process more complicated. In the unsupervised case, the canonical intent permutation is the identity ordering $(1,2,...,K)$ as the intent numbers are completely symmetric and not linked to any meaningful intent label. However, the true intent labels are already known in the supervised case. Then a challenging problem is to determine the canonical permutation $\bigpi_0$. Moreover, for the part of labeled documents, we have the true intent structure $\bigz_d$, which is unnecessary to draw from the GMM-Multi prior. So the next challenging problem is how to leverage the known intent structure $\bigz_d$ to help for learning. That is, how to obtain $\bigu_d$ and $\bigpi_d$ using $\bigz_d$. Below, we discuss in detail on how to solve these problems.

\begin{figure}[t]
\centering
\includegraphics[height=1.50in, width=2.7in]{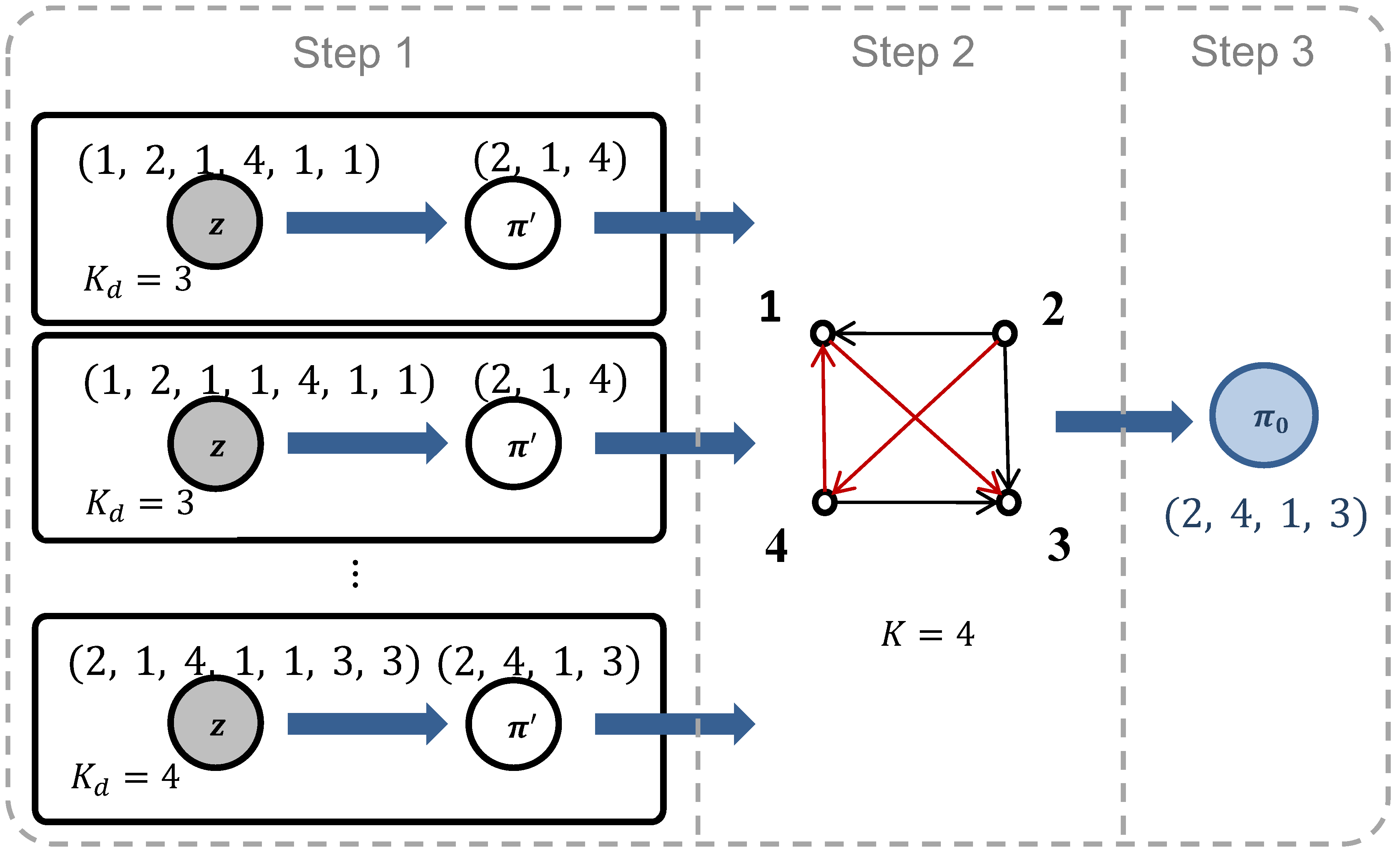}
\caption{The graphical expression of computing $\bigpi_0$.}
\label{fig:pi}
\end{figure}

\textbf{For $\!\bigpi_0\!\!$ : } As shown in Fig.\ref{fig:pi}, we present an approximate three-step algorithm to obtain the canonical permutation $\bigpi_0$. {\bf{Step 1:}} We compute $\bigpi'_d$ for each labeled document $\bigs_d$, where $\bigpi'_d$ is the permutation of the $K_d$ ($\leqslant K$) intent labels appearing in document $\bigs_d$. We arrange the $K_d$ intent labels in the order they appear in $\bigz_d$. However, there exist cases where the same intent label appears in disconnected portions of $\bigz_d$, which are rare in practice. When encountering these cases, we take the position that the label appears most consecutively in the sequence. If there are two or more of these occurrences, we take the first position (see Fig.~\ref{fig:pi} for some examples). {\bf{Step 2:}} We introduce variables $g_{ij}(i,j\in [K])$, where $g_{ij}$ counts the times of the intent label $i$ appearing before $j$ in all $\bigpi'_d$. Then we define a directed graph $\mathcal{G}=(\mathcal{V}, \mathcal{E})$, where $\mathcal{V}\in [K]$ is the set of nodes and $\mathcal{E}=\{(i, j)|g_{ij}>=g_{ji};i,j\in \mathcal{V}\}$ is the set of edges. {\bf{Step 3:}} We obtain the $\bigpi_0$ by calculating the topological sequence of $\mathcal{G}$. If there are circles in $\mathcal{G}$, we randomly break them. If there are multiple topological sequences, we randomly take one. In our experiments, the real situations are always in accordance with our intuition that only one topological sequence can be obtained from $\mathcal{G}$.

\textbf{For $\bigu_d$ and $\bigpi_d$ : } To combine the label information, we compute $\bigu_d$ and $\bigpi_d$ using $\bigz_d$ and $\bigpi_0$. $\bigu_d$  can be obtained by directly putting all the ordered elements of $\bigz_d$ into a bag. $\bigpi_d$ is a permutation of $K$ numbers, which can be obtained by inserting the remaining $K-K_d$ numbers into $\bigpi'_d$ . We want $\bigpi_d$ to be as close to $\bigpi_0$ as possible, which is consistent with the idea of GMM. Specially, $d(\bigpi_d, \bigpi_0)$ is the distance from $\bigpi_d$ to $\bigpi_0$, which defined as the minimum number of swaps of adjacent elements needed to transform $\bigpi_d$ into the same order of $\bigpi_0$. Inspired by the idea of greedy method, we insert the  $K-K_d$ numbers into $\bigpi'_d$ one by one and each step need to minimize the distance. During supervised learning, when it comes to a labeled document $\bigs_d$, we directly take the values of $\bigpi_d$ and $\bigu_d$ instead of updating them.

\begin{table*}[!t]
\centering
\caption{\label{tab_clustering}Results of unsupervised clustering.}
\scalebox{.81}{
\begin{tabu}{|X[0.3l]|X[2.7l]|X[1c]|X[1c]|X[1c]|X[1c]|X[1c]|X[1c]|X[1c]|X[1c]|}
\hline
\multicolumn{2}{|c|}{\multirow{2}{*}{{\bf Models}}} & \multicolumn{4}{|c|}{\bf Chemical  }& \multicolumn{4}{|c|}{\bf Elements}\\ \cline{3-10}
\multicolumn{2}{|c|}{ } & \bf ARI & \bf Recall & \bf Prec & \bf Fscore & \bf ARI & \bf Recall & \bf Prec & \bf Fscore \\ \hline
\multirow{6}{*}{\rotatebox{90}{$K=5$}}
&    K-means            &$0.0023$&$0.7139$&$0.4257$&$0.5333$&$0.0236$&$\textbf{0.8107}$&$0.3201$&$0.4583$\\
&    Boilerplate-LDA    &$0.3953$&$0.6797$&$0.6647$&$0.6720$&$0.3060$&$0.6975$&$0.5091$&$0.5884$\\
&    GMM                &$0.1942$&$0.5093$&$0.5982$&$0.5501$&$0.3395$&$0.7004$&$0.5348$&$0.6064$\\
&    GMM-LDA (Uniform)  &$0.2614$&$0.6886$&$0.5836$&$0.6314$&$0.1911$&$0.6002$&$0.4449$&$0.5106$\\
&    GMM-LDA            &$0.4055$&$0.7473$&$0.6699$&$0.7065$&$0.3223$&$0.7176$&$0.5271$&$0.6075$\\
&    EGMM-LDA           &$\textbf{0.4125}$&$\textbf{0.7553}$&$\textbf{0.6725}$&$\textbf{0.7115}$&$\textbf{0.3567}$&$0.7450$&$\textbf{0.5517}$&$\textbf{0.6339}$\\
\hline
\multirow{6}{*}{\rotatebox{90}{$K=10$}}
&    K-means             &$0.0011$&$0.5779$&$0.4332$&$0.4949$&$0.0504$&$\textbf{0.6895}$&$0.3460$&$0.4606$\\
&    Boilerplate-LDA     &$0.2861$&$0.5367$&$0.6955$&$0.6056$&$0.3511$&$0.5885$&$0.6440$&$0.6149$\\
&    GMM                 &$0.1381$&$0.3302$&$0.6268$&$0.4324$&$0.3652$&$0.5692$&$0.6491$&$0.6063$\\
&    GMM-LDA (Uniform)   &$0.3206$&$0.6069$&$0.6227$&$0.6138$&$0.3432$&$0.5720$&$0.6146$&$0.5924$\\
&    GMM-LDA             &$0.4477$&$0.6747$&$0.7311$&$0.7017$&$0.4013$&$0.6273$&$0.6857$&$0.6549$\\
&    EGMM-LDA            &$\textbf{0.4546}$&$\textbf{0.6786}$&$\textbf{0.7364}$&$\textbf{0.7063}$&$\textbf{0.4196}$&$0.6418$&$\textbf{0.6865}$&$\textbf{0.6631}$\\
\hline
\end{tabu}}
\end{table*}

\section{GMM-LDA with Entropic Regularization}

GMM-LDA jointly models the two incompatible structures of documents by using a binary variable to indicate the type (intent or topic) of each word. It can happen that a same word located in two different positions are assigned with different types, which is somewhat unreasonable. Most of the time, the type of a word can be decided regardless of which position it appears in. For instance, the words ``propose'' and ``experiment'' are more likely to be intent words regardless of their positions. In order to model the significant divergence between topics and intents, we introduce the entropy of the words to make our model more descriptive. As we know, in information theory, the entropy of a discrete random variable $X=\{x_1, x_2, ..., x_n\}$ can explicitly be written as $\mathrm{H}(X)=-\sum_{i=1}^n p(x_i)\log p(x_i)$. Similarly, the entropy of a word $w_v$ in the vocabulary can be formulated as:
\begin{eqnarray}
\mathrm{H}(b_v)=-\sum_{i=0,1} p(b_v=i)\log p(b_v=i),
\end{eqnarray}where $p(b_v=i)$ denotes the probability that word $w_v$ appears as an intent word ($i=0$) or an topic word ($i=1$) in the corpus. Lower entropic value means better separation.

As GMM-LDA is under Bayesian inference framework, it is challengeable to incorporate the entropic knowledge. Nevertheless, regularized Bayesian inference framework \cite{zhu2014bayesian} provides us an alternative interpretation of Bayesian inference \cite{williams1980bayesian} and can combine domain knowledge flexibly. Specifically, GMM-LDA with entropic regularization can be formulated as:
\begin{eqnarray}\label{ent}
\min_{q(\mathcal{M})\in \mathcal{P}} { \mathrm{KL}} (q(\mathcal{M})||q(\mathcal{M}|\mathcal{D})) + c \sum_{v=1}^V \mathrm{H} (b_v),
\end{eqnarray}where $\mathcal{P}$ is the space of probability distributions and  $c\geqslant 0$ is a regularization parameter. When $c=0$, the optimal solution of the Kullback-Leibler divergence ${ \mathrm{KL}} (q||p)$ is $q(\mathcal{M}) = q(\mathcal{M}|\mathcal{D})$, the standard Bayesian posterior distribution as in Eq. (\ref{bayes}). When $c>0$, the entropic knowledge is imposed as a regularization constraint. With mean-filed assumption, the inference of Eq. (\ref{ent}) is similar to that of GMM-LDA. Moreover, the entropic regularization is only relevant to $\bigb$, which can be combined with both the unsupervised GMM-LDA and the supervised GMM-LDA.

\begin{table}[t]
\centering
\caption{\label{tab_corpus}Statistics of datasets.}
\scalebox{.88}{
\begin{tabular}{|l|c|c|c|c|}
\hline \bf Corpus  & \bf Docs & \bf Sentences &  \bf Vocab &\bf Tokens \\ \hline
Chemical & $965$ & $9,488$  & $4,981$ & $123,119$\\\hline
Elements  & $118$ & $1,848$ &  $5,301$ & $67,484$\\
\hline
\end{tabular}}
\end{table}

\section{Experimental Results}

To demonstrate the efficacy of our models, we evaluate the performance on two tasks: unsupervised clustering and supervised classification.
We use two real datasets:
1) {\bf{Chemical}}~\cite{guo2010identifying}: It contains $965$ abstracts of scientific papers about $5$ kinds of chemicals, and each abstract focuses on one of the $5$ topics. Each sentence is annotated with one of the $7$ intent labels: \emph{Background}, \emph{Objective}, \emph{Related Work}, \emph{Method}, \emph{Result}, \emph{Conclusion} and \emph{Future Work};
and 2) {\bf{Elements}}~\cite{Chen09}: It consists of $118$ articles from the English Wikipedia, and each article talks about one of the $118$ chemical elements in the periodic table. Each paragraph is annotated with an intent label. We take the $8$ most frequently occurring intent labels:  \emph{Top-level Segment}, \emph{History}, \emph{Isotopes}, \emph{Applications}, \emph{Occurrence}, \emph{Notable Characteristics}, \emph{Precautions} and \emph{Compounds}, and filter out paragraphs with other labels. Although the intent structure is paragraph-level in Element, while it is sentence-level in Chemical, the word ``sentence'' is used throughout the paper for simplicity. Tab.~\ref{tab_corpus} summarizes the dataset statistics. Although both datasets are in chemistry domain, they have different characteristics that can be observed from the experimental results.
In Chemical, the intent orders are relatively fixed due to the writing conventions of scientific papers, while they are more variable in Elements.

Data preprocessing involves removing a small set of stop words, tokens containing non-alphabetic characters, tokens appearing less than $3$ times, tokens of length one and sentences with less than $5$ valid tokens. We report the average results over $5$ runs, while each run takes a sufficiently large number of iterations (e.g. $2000$) to converge. Statistical significance is measured with \emph{t-test}.

\subsection{Unsupervised Clustering}

Our goal of unsupervised clustering is to learn an intent label for each sentence in the corpus without any true label information. Adjusted Rand Index (ARI)\cite{vinh2010information}, recall, precision and F-score are used as our evaluation measures. F-score is the harmonic mean of recall and precision. For all the four measures, higher scores are better.

We consider two variants of our model:
1) {\bf GMM-LDA}: Our unsupervised model;
and 2) {\bf EGMM-LDA}: GMM-LDA with entropic regularization. For hyperparameters, we set $\theta_0=0.1$, $\lambda_0=0.1$, $\alpha_0=0.1$, $\beta_0=0.1$ and $\gamma_0=1$, since we find that the results are insensitive to them. $\nu_0$ is set to be $0.1$ times the number of documents in the corpus. For EGMM-LDA, we set the regularization parameter $c$ to be $0.1$.
The baseline methods we use are:
1) {\bf K-means:} The feature used for each sentence is the bag of words.
2) {\bf Boilerplate-LDA}: The model presented in~\cite{rhetorical14}.
3) {\bf GMM}: The intent part of GMM-LDA, which is the content model by \cite{Chen09} and can be implemented by fixing all indicator variables $b_{dsm}$ to $0$ during learning;
and 4) {\bf GMM-LDA (Uniform)}: The model assumes a uniform distribution over all intent permutations and can be implemented by fixing $\bigrho$ to zero.

{\bf Clustering performance:}  We set the number of topics $T=10$ for Chemical and $T=5$ for Elements. Tab.~\ref{tab_clustering} shows the results. For Chemical, our two models outperform all other methods on four measures, with $p$-values smaller than $0.001$ except for Boilerplate-LDA with $K=5$. The exception of Boilerplate-LDA may be caused by the large variance in the results of multiple runs when $K$ is small. Since the first-order Markov chain is used in Boilerplate-LDA for order structure learning, which only has a local view and is more susceptible to noise. Moreover, first-order Markov chain would select the same intent label for disconnected sentences within a document, which is against our intuition. Our models overcome these problems by using GMM and also achieve better performance. The simpler variants of our models achieve reasonable performance. GMM underperforms GMM-LDA, indicating that modeling topics and intents simultaneously provides a richer and more effective way to document structure learning. The bad performance of the uniform variant proves the indispensability to model the intent order structure. Moreover, our model yields better results with entropic regularization. As to Elements, the results are similar to that of Chemical. However, we can observe that GMM performs competitively in Elements, which shows the characteristic of this dataset that there is no obvious topic structure. We can also observe that the highest recall scores are obtained by K-means with very low precision scores, since K-means prefers to assign the same label to most of the sentences.
It can thus be seen that the task is difficult and the richer models are required.

\begin{figure}[!t]
\centering
\includegraphics[height=1.5in, width=2in]{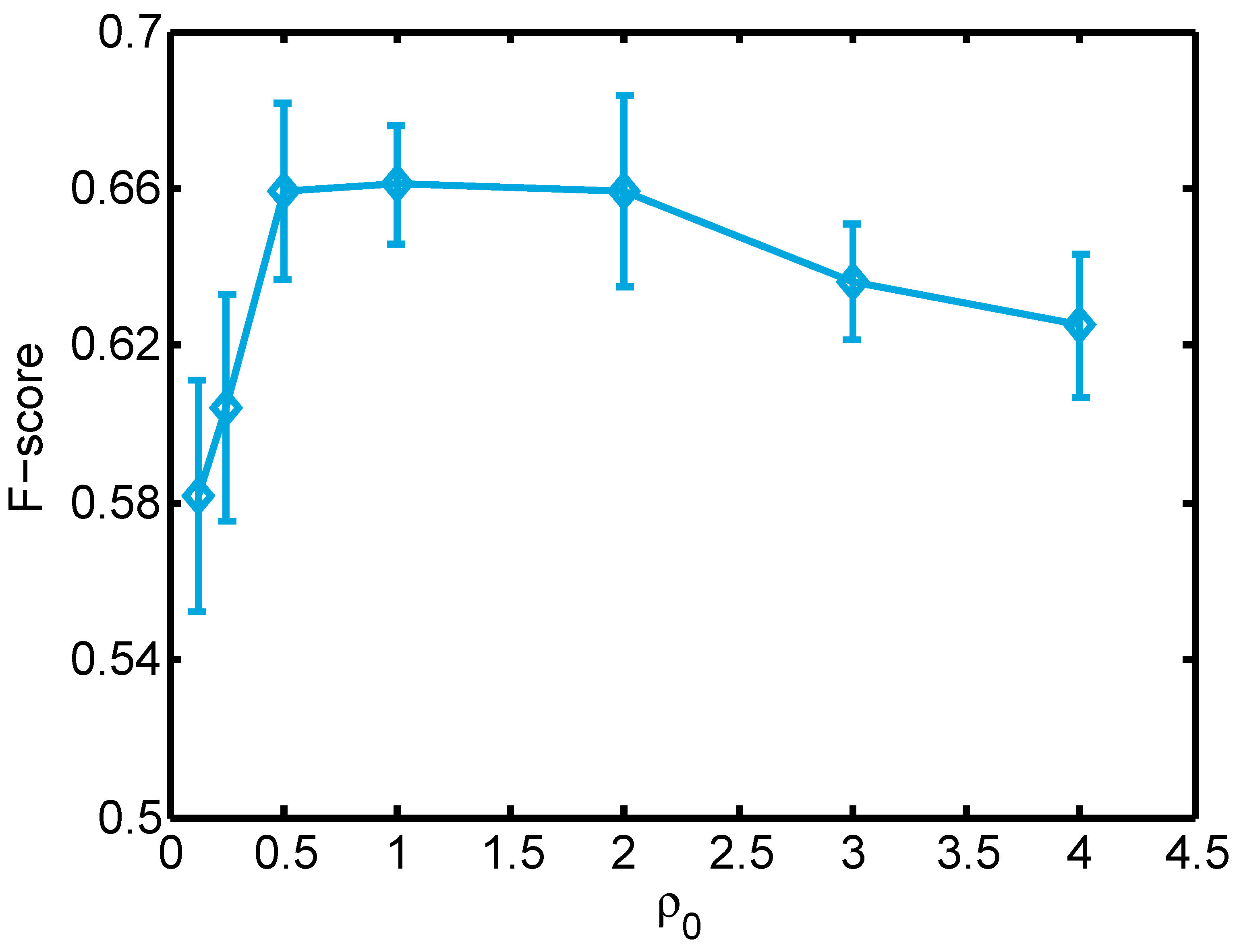}
\caption{Sensitivity of $\rho_0$.}
\label{tunerho}
\end{figure}

{\bf Hyperparameter $\rho_0$:} $\rho_0$ controls the variability of the order structure and would be set according to different datasets. The model with large $\rho_0$ assigns massive probabilities around the canonical permutation and the order structure is relatively fixed, while the model with small $\rho_0$ relaxes the constraints and has a variety of orders far from the canonical one.  For Elements with $K=10$, we change $\rho_0$ from $0.125$ to $4$ and report the F-scores of GMM-LDA in Fig.~\ref{tunerho}. It can be observed that the performance is stable in a wide range (e.g. $0.5<\rho_0<2$). We set $\rho_0=2$ for all the experiments except for Elements with $K=10$, in which we set $\rho_0=1$.

\begin{table}[h]
\centering
\caption{\label{wordsytype}Most commonly words of two types on Chemical.}
\scalebox{.81}{
\begin{tabular}{|lll|lll|}
\hline
\multicolumn{3}{|c|}{\bf Intent Words} & \multicolumn{3}{|c|}{\bf Topic Words  }\\ \hline
~days &~results &~increased~&~rats &~protein &~expression~\\
~other  &~control &~observed~&~mice&~levels&~exposure~\\
~used &~present&~significant~&~DNA &~activity &~adducts~\\
~more&~treated&~between~&~gene&~effects &~treatment~\\
~data&~suggest &~compared~&~cells &~tumor &~phenobarbital~\\
~study &~human &~showed~&~liver&~induced &~formation~\\                   \hline
\end{tabular}}
\end{table}

{\bf Types of words:} To embody the intents and topics in the results, we assume that each word in the vocabulary is either an intent word or a topic word. If a word appears in the corpus more as an intent word than a topic word, we classify it as an intent word; otherwise, it is a topic word. Note that this additional condition is introduced only for the ease of demonstration. In order to see how our models separate these two types, we list $18$ most commonly used words of each type in Chemical according to the result of EGMM-LDA. As shown in Tab. \ref{wordsytype}, we can observe that almost all the intent words has rhetorical functions that can express the intents of sentences, while almost all the topic words are about chemical topics.
The good separations justify our assumption and show the effectiveness of our models.

\begin{table}[!t]
\centering
\caption{\label{tab_class}Results of supervised classification.}
\scalebox{.81}{
\begin{tabular}{|l|c|c|c|c|}
\hline
\multicolumn{1}{|c|}{\multirow{2}{*}{{\bf Models}}} & \multicolumn{2}{|c|}{\bf Chemical }& \multicolumn{2}{|c|}{\bf Elements}\\ \cline{2-5}
& \bf ARI & \bf ACC & \bf ARI & \bf ACC\\ \hline
SVM                 &$0.399$&$0.674$&$0.326$&$0.622$\\
sBoilerplate-LDA    &$0.506$&$0.709$&$0.522$&$0.749$\\
sGMM                &$0.276$&$0.530$&$0.531$&$0.744$\\
sGMM-LDA            &$\textbf{0.510}$&$\textbf{0.731}$&$0.521$&$0.752$\\
sEGMM-LDA           &$\textbf{0.510}$&$\textbf{0.730}$&$\textbf{0.562}$&$\textbf{0.775}$\\
\hline
\end{tabular}}
\end{table}

\subsection{Supervised Classification}
Now, we evaluate our supervised models for classifying sentences. For each dataset, we randomly choose $20\%$ documents; annotate their sentences with intent labels; and use them for training. Our goal is to learn the intent labels for the sentences in the remaining $80\%$ documents. We report accuracy (ACC) and the ARI scores to show the improvements compared to the unsupervised learning. We again consider two variants of our model:
1) {\bf sGMM-LDA}: Our supervised model;
and 2) {\bf sEGMM-LDA}: sGMM-LDA with entropic regularization.
The baseline methods are :
1) {\bf SVM:} We use the bag-of-words features, linear kernel and SVM-Light tools~\cite{joachims1999making}.
2) {\bf sBoilerplate-LDA :} The supervised version of Boilerplate-LDA, in which we fix the known labels during learning instead of updating them;
and 3) {\bf sGMM :} The intent part of our sGMM-LDA. All the settings are the same as that in the unsupervised learning.

{\bf Classification performance:} Tab. \ref{tab_class} presents the results. For Chemical, the best accuracies are achieved by our two models ($p<0.01$), which again proves that our assumption of the two types of words is reasonable and the intent order structure can be better modeled by employing GMM. For Elements, sBoilerplate-LDA and sGMM perform competitively, while EGMM-LDA beats all the other methods ($p<0.05$). It shows that our model is more robust with entropic regularization. The ARI scores improve a lot compared to that in Tab. \ref{tab_clustering}, indicating that the predictive power can be largely improved with just $20\%$ labeled documents.

\begin{table}[h]
\centering
\caption{Results of the canonical intent permutation and the intent words on Chemical with sEGMM-LDA. }
\scalebox{.8}{
\begin{tabular}
{|c|l|l|}
\hline
\bf No. & \multicolumn{1}{|c|}{\bf Intent Label} &\multicolumn{1}{|c|}{ \bf High-Frequency Words} \\ \hline
0 & \emph{Background} & cancer, studies, carcinogen, environmental,\\
&& aromatic, known, shown, however, used  \\
1 & \emph{Objective} & study, using, investigated, present, examined,\\
&& used, whether, determine, effect, evaluated\\
2 & \emph{Method} &    days, treated, diet, single, control, groups, \\
&& body, followed, water, using, injection\\
3 & \emph{Result} &  increased, significantly, observed, compared,\\
&& showed, higher, however, respectively\\
4 & \emph{Related Work} & polymerase, significantly, results, codes, ICD, \\
&& comparisons, percent, suggest, enhances\\
5 & \emph{Conclusion} & results, suggest, indicate, findings, study, role,\\
&& studies, thus, conclusion, important\\
6 & \emph{Future Work} & needed, however, needs, required, future, \\
&&research, investigations, studies, observed\\ \hline
\end{tabular}}
\label{intentword}
\end{table}

{\bf Intent words:} We can obtain the canonical intent permutation by the known intent labels, at the same time the distribution over vocabulary can be learned. Tab. \ref{intentword} shows the results on Chemical with sEGMM-LDA. We can observe that the canonical intent order (numbered from $0$ to $6$) conforms to the convention in scientific writing. Moreover, from the high-frequency words of each intent, we can see that most of the words express the intent labels well. For instance, ``study'' and ``investigated'' express the intent \emph{Objective}, while ``increased'' and ``significantly'' are for \emph{Result}.

\section{Related Work}
From the algorithmic perspective, our work is grounded in topic models, such as Latent Dirichlet Allocation (LDA) \cite{blei2003}, which have been widely developed for many NLP tasks. Instead of representing documents as bags of words, many expanded models take specific structural constraints into consideration \cite{purver2006unsupervised,gruber2007hidden}. Among different models, our work has a closer relation to the models with order structure. For order modeling, Markov chain can only capture the dependence locally \cite{rhetorical14,barzilay2004catching,elsner2007unified}, while the generalized Mallows model (GMM) \cite{fligner1986distance} has a global view \cite{Chen09,du2015topic,cheng2009decision}. A more complete model can be obtained by dividing the words into different types. In early trial of zoneLDAb \cite{varga2012}, a type of words are for describing background, which are independent of the category of the sentence. Boilerplate-LDA \cite{rhetorical14} also considers two types: document-specific topic words and rhetorical words. Three types of words are learnt by a rule-based method in \cite{nguyen2015topic}. However, global order structure is not considered in these models. Therefore, jointly modeling topics and intents with global order structure is of great value.

\section{Conclusion and Future Work}
We present GMM-LDA (both unsupervised and supervised) for document structure learning, which simultaneously model topics and intents. The generalized Mallows model is employed to model the intent order globally. Moreover, we consider the entropic regularization to make the model more descriptive. Our results demonstrate the reasonability of our intuitions and the effectiveness of our models. For future work, we are interested in making our models richer by combining local coherence constraints.

\section{ Acknowledgments}
This work was mainly done when the first author was an intern at Microsoft Research Asia. The work was supported by the National Basic Research Program (973 Program) of China (Nos. 2013CB329403, 2012CB316301), National NSF of China (Nos. 61322308, 61332007), TNList Big Data Initiative, and Tsinghua Initiative Scientific Research Program (Nos. 20121088071, 20141080934).

\bibliographystyle{aaai}
{\small{
\bibliography{chen-zhu}}}
\end{document}